\documentclass[a4paper,twoside]{article}

\usepackage{epsfig}
\usepackage{subcaption}
\usepackage{calc}
\usepackage{graphicx}
\usepackage{times}
\usepackage{comment}
\usepackage{amssymb}
\usepackage{amstext}
\usepackage{amsmath}
\usepackage{amsthm}
\usepackage{multicol}
\usepackage{pslatex}
\usepackage{apalike}
\usepackage{xspace}
\usepackage{mathtools}
\usepackage{algorithm, algorithmic}
\usepackage{setspace}

\usepackage{SCITEPRESS}     % Please add other packages that you may need BEFORE the SCITEPRESS.sty package.

\DeclarePairedDelimiter\floor{\lfloor}{\rfloor}

\newcommand{\etal}{\textit{et al}.\xspace}

\usepackage[breaklinks=true,letterpaper=true,colorlinks,bookmarks=false]{hyperref}

\begin{document}

\title{A CNN-Based Feature Space for Semi-Supervised Incremental Learning\\ in Assisted Living Applications}

\author{\authorname{Tobias Scheck\orcidAuthor{0000-0002-1829-0996}, Ana Perez Grassi\orcidAuthor{0000-0003-1171-903X}, Gangolf Hirtz\orcidAuthor{0000-0002-4393-5354}}
    \affiliation{Faculty of Electrical Engineering and Information Technology, Chemnitz University of Technology, Germany}
    \email{\{tobias.scheck,ana-cecilia.perez-grassi,g.hirtz\}@etit.tu-chemnitz.de}
}


%%%%%%%%% ABSTRACT
\abstract{
    A Convolutional Neural Network (CNN) is sometimes confronted with objects of changing appearance ( new instances) that exceed its generalization capability. This requires the CNN to incorporate new knowledge, i.e., to learn incrementally. In this paper, we are concerned with this problem in the context of assisted living. We propose using the feature space that results from the training dataset to automatically label problematic images that could not be properly recognized by the CNN. The idea is to exploit the extra information in the feature space for a semi-supervised labeling and to employ problematic images to improve the CNN's classification model. Among other benefits, the resulting semi-supervised incremental learning process allows improving the classification accuracy of new instances by $40\%$ as illustrated by extensive experiments.
}

\onecolumn \maketitle \normalsize \setcounter{footnote}{0} \vfill

%%%%%%%%% BODY TEXT
\section{\uppercase{Introduction}}

\noindent Convolutional Neural Networks (CNNs) are used for all kinds of object recognition/classification based on images. The basic idea is that CNNs {\em learn} how to distinguish objects of interest from labeled images. Although the ultimate goal of object classification is to identify any possible object of any possible category or class, for real-world applications, it is usually not feasible to generate a sufficient number of labeled images. Popular datasets such as MS COCO and ILSVRC \cite{fleet_microsoft_2014,russakovsky_imagenet_2015} contain 80 and 1000 classes respectively and, hence, are not sufficient to describe all possible objects \cite{han_advanced_2018}.

On the other hand, most applications are not concerned with detecting all kinds of objects, but only a small subset of them related to their tasks/objectives. For example, applications in the automotive domain are typically focused on recognizing vehicles, traffic signs, pedestrians, cyclists, etc., while other objects like those carried by pedestrians are not relevant. Similarly, assisted living applications are concerned with daily-life objects like mugs, chairs, tables, etc., while recognizing cars or traffic signs is out of scope.

This restriction to few object classes allows optimizing CNNs for a specific task. However, a real-life environment still undergoes continuous change. As a result, CNNs need to incorporate new knowledge, i.e., implement lifelong/incremental learning \cite{chen_fine-tuning_2017,parisi_continual_2019}, which is a challenging endeavor bearing the risk of {\em catastrophic forgetting} \cite{goodfellow_empirical_2013}, i.e., losing the ability to recognize known objects.

\textbf{Contributions. }In this work, we are concerned with the above problem in assisted living applications. The user in this context makes repeated use of the same objects, i.e., the same mug, the same chair, etc., which are thus easy to recognize with a CNN. On the other hand, daily-life objects are often replaced by new ones that, although belonging to the same class, can have a very different appearance, i.e., new instances. This sometimes exceeds the CNN's capability of generalizing, which stops detecting these images reliably.

We propose a technique that combines semi-supervised labeling and incremental learning to approach a personalized assisted living system, capable of adapting to new instances introduced by the user at any point in time. To this end, an acquisition function selects and stores problematic images that could not be classified with a satisfactory level of confidence. We then make use of the feature space generated from the training dataset to label these images without human intervention. Since the feature space contains more information than the CNN's classification model, it allows reliably classifying new instances with only a small amount of label noise. Finally, the labeled problematic images are incorporated into the CNN's classification model by fine-tuning. %We perform extensive evaluations showing the benefits of

\noindent\textbf{Structure of the paper.} This paper is organized as follows. Section \ref{sec:related} introduces the state of the art, whereas the proposed approach is described in Sec.~\ref{sec:system}. Section \ref{sec:experimet} then evaluates the proposed approach and Sec.~\ref{Sec:Conclusion} concludes the paper.%This

\section{RELATED WORK}
\label{sec:related}

\noindent There is an increasing interest in techniques such as lifelong/incremental learning and semi-supervised labeling, which aim to alleviate CNNs' dependency on huge amounts of labeled data. While incremental learning focuses on the capability to successively learn from small amounts of data, semi-supervised labeling looks for methods to replace the usually expensive and time-consuming labeling of datasets.

An overview of incremental learning techniques is presented by Parisi \etal \cite{parisi_continual_2019}. One challenge of incremental learning is to avoid catastrophic forgetting \cite{mccloskey_catastrophic_1989}. That refers to the problem of new learning interfering with old learning when the network is trained gradually. An evaluation of catastrophic forgetting on modern neural networks was presented by Goodfellow \etal in \cite{goodfellow_empirical_2013}. Further, new metrics and benchmarks for measuring catastrophic forgetting are introduced by Kemker \etal \cite{kemker_measuring_2017}.

Although retraining from scratch can prevent catastrophic forgetting from happening, this is very inefficient. Approaches to mitigate catastrophic forgetting are typically based on {\em rehearsal}, {\em architecture} and/or {\em regularization} strategies. Rehearsal methods interleave old data with new data to fine-tune the network \cite{rebuffi_icarl:_2016}. In \cite{hayes_memory_2018}, Hayes \etal study full rehearsal (i.e., involving all old data) in deep neural networks. In this work, we also use a mix of old and new data, however, similar to \cite{chen_fine-tuning_2017} we only use a small percentage of old data, i.e., {\em partial} rehearsal. Architecture methods use different aspects of the network's structure to reduce catastrophic forgetting \cite{rusu_progressive_2016,lomonaco_core50:_2017}. Further, regularization strategies \cite{li_learning_2016,kirkpatrick_overcoming_2016} focus on the loss function, which is modified to retain old data, while incorporating new one. These alleviate catastrophic forgetting by limiting how much neural weights can change. Basic regularization techniques include weight sparsification, dropout and early stopping. Further works combine regularization with architecture methods \cite{maltoni_continuous_2018}, as well as with rehearsal methods \cite{rebuffi_icarl:_2016}. In this paper, we opt to combine partial rehearsal with early stopping, since this better suits our application an provides good results.

In \cite{chen_fine-tuning_2017}, Käding \etal conclude that incremental learning can be directly achieved by continuous fine-tuning. Our paper is in line with this work, however, in contrast to \cite{chen_fine-tuning_2017}, the new data added during each incremental learning step may belong to different classes reflecting the nature of assisted living applications.

The concept of {\em active learning} \cite{gal_deep_2017} also allows counteracting CNNs' dependency on labeled data. Active learning implies first training a model with a relatively small amount of data and only letting an {\em oracle} --- often a human expert --- label further data to retrain the model, if they are selected by an acquisition function. This process is then repeated with the training set increasing in size over time. As already mentioned, we propose replacing the oracle by a semi-supervised process, which labels the selected data using the feature space generated from the training dataset.

With respect to semi-supervised labeling, Lee \cite{dong-hyun_lee_pseudo-label_2013} proposed assigning pseudo-labels to unlabeled data selecting the class with the highest predicted probability. In \cite{enguehard_semi-supervised_2019}, Enguehard \etal present a semi-supervised method based on embedded clustering, whereas Rasmus \etal propose combining a Ladder network with supervised learning in \cite{rasmus_semi-supervised_2015}.

\section{\uppercase{System description}}
\label{sec:system}

\begin{figure*}[h!]
    \centering
    \includegraphics[width=\textwidth]{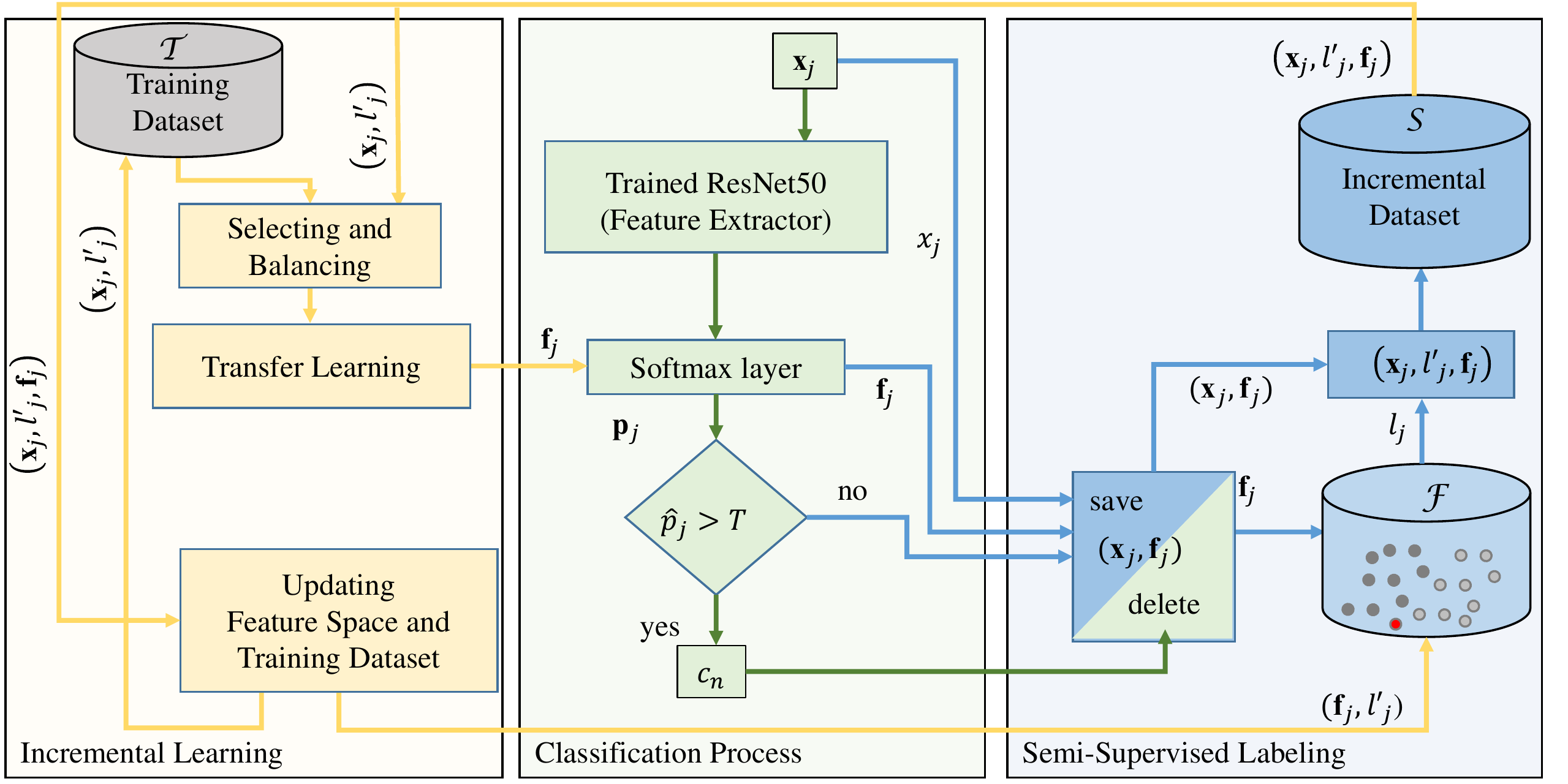}
    \caption{Proposed semi-supervised incremental learning system for assisted living application.}
    \label{fig:System}
\end{figure*}

\noindent As shown in Fig.~\ref{fig:System}, our system can be divided in three processes: classification, semi-supervised labeling and incremental learning. The {\em classification} process is based on a trained CNN and performs the main task of the system. It takes an image and assigns it a class according to a computed confidence value. During this process an acquisition function selects those images with unsatisfactory classification results (i.e., with confidence value lower than a given threshold) and forwards them, together with their {\em feature vectors}, to the semi-supervised labeling process.

The {\em semi-supervised labeling} process then tags these images according to a pre-stored feature space. This feature space is generated from the training data and then successively updated during the incremental learning process. The resulting labels together with their images are incorporated into the CNN's classification model by fine-tuning during the  {\em incremental learning} process. Finally, the new labeled images are further added to the training dataset, their feature vectors are added to the feature space and the classification model is updated.

Note that copies of the classification model, the training dataset and its corresponding feature space need to be stored for the semi-supervised labeling and the incremental learning processes. However, this data is only required offline and does not affect performance, albeit increasing memory demand. In the following sections we describe each process in detail.
\subsection{Classification Process}
\label{sec:classificationProcess}

The classification process is responsible for identifying objects displayed on the input images. This process is the only one that runs online and is visible to the user. The classification is performed by a pre-trained CNN. In this work, we use the well-established ResNet50 \cite{he_deep_2015}, which not only acts as a classification network, but also as feature extractor. ResNet50 can be described as a feature extractor followed by a fully connected Softmax layer, where the first one generates feature vectors and the second one classifies them.

Let $\mathcal{C}=\{c_n\}$ with $1\leq n \leq N$ be the set of $N$ object classes considered by the system. The training dataset used to generate the first model $\mathcal{M}_0$ is denoted by $\mathcal{T}_0=\{(\mathbf{x}_i, l_i)\,\,|\,\,l_i \in \mathcal{C},\,1\leq i \leq |\mathcal{T}_0|\}$, where $\mathbf{x}_i$ is an image and $l_i$ its corresponding label.%, $l_i \in \mathcal{C}$ and $1\leq i \leq |\mathcal{T}_0|$.

Once the first model $\mathcal{M}_0$ is generated, $\mathcal{T}_0$ is passed through the network in order to obtain its feature space. ResNet50  generates for each image $\mathbf{x}_i$ a feature vector $\mathbf{f}_i:=f(\mathbf{x}_i)$ of 2048 elements \cite{he_deep_2015}. The set of all feature vectors from $\mathcal{T}_0$ together with their corresponding labels $l_i$ is denoted $\mathcal{F}_0=\{(\mathbf{f}_i,l_i)\}$ and constitutes our first feature space. Finally, $\mathcal {T}_0$ and $\mathcal{F}_0$ are stored to be used during the semi-supervised labeling and incremental learning processes.

During the classification process, ResNet50 generates a feature vector $\mathbf{f}_j$ for each input image $\mathbf{x}_j$. This vector is then passed to the final fully connected Softmax layer, which returns a vector $\mathbf{p}_j=[p_j^1,\cdots,p_j^N]$, where $p_j^n$ denotes the probability of $\mathbf{x}_j$ to belong to class $c_n$. Finally, $\mathbf{x}_j$ is classified according to the greatest probability $\hat{p}_j\in\mathbf{p}_j$, where $\hat{p}_j:=\max\{p^1_j,\cdots, p^N_j\}$. This means that $\mathbf{x}_j$ is assigned to the class $c_n$, for which $\hat{p}_j=p^n_j$ holds.

The value of $\hat{p}_j$, called confidence value, gives an idea of how sure the classification model is about the class assigned to the image $\mathbf{x}_j$. Therefore, when an image $\mathbf{x}_j$ is classified with a confidence value below a certain threshold $t$, we conclude that the model is not sufficiently sure about the nature of the imaged object and its classification is considered invalid.

Images that are classified with a low confidence value constitute a valuable source of knowledge for the system. These contain information about objects of interest, which is not considered in the current classification model $\mathcal{M}_q$, with $q\in \mathbb{N}_0$. In order to learn from these images later, an acquisition function $f_a(\hat{p}_j,t)$ is defined based on the confidence value and a given threshold:
\begin{eqnarray}
    \label{eq:Acquisition}
    f_a(\hat{p}_j,t)
    \begin{cases}
        (\mathbf{x}_j,\mathbf{f}_j)\text{ selected}  & \text{if }\hat{p}_j< t, \\
        (\mathbf{x}_j,\mathbf{f}_j)\text{ discarded} & \text{else}.
    \end{cases}
\end{eqnarray}
The selected images $\mathbf{x}_j$ together with their feature vectors $\mathbf{f}_j$ are then passed to the labeling process.% discussed next.

\subsection{Labeling Process}
\label{sec:Labeling_Process}

\noindent The images selected by $f_a(\hat{p}_j,t)$ contain objects whose class could not be satisfactorily identified. That is, either objects belong to a new class not included in $\mathcal{C}$, or they belong to a known class, but are not properly represented by the images in the current training dataset $\mathcal{T}_q$. In this paper we focus on the latter, since this is the most common case in the application of interest.

As mentioned before, ResNet50 is constituted by a feature extractor followed by a Softmax layer \cite{he_deep_2015}. During training, both, the feature extractor and the Softmax layer adapt their weights iteratively according to a training dataset and a loss function. At the end of training, all weights inside the feature extractor and the Softmax layer are fixed, determining how to compute the features and how to assign them a class. This means that the information about how to separate classes inside a feature space $\mathcal{F}_q$ is summarized in the weights of the Softmax layer.

The Softmax layer does not memorize the complete feature space, on the contrary, it learns a representation of it, which should be sufficiently precise to distinguish between classes and, at the same time, general enough not to overfit. This results in an efficient classification method, but it also implies loss of information. In particular, this loss of information affects those images, that are underrepresented in the training dataset. In many cases, although the feature space representation learned by the Softmax layer does not describe these images correctly, the complete feature space still does. As a consequence, the feature space is suitable for labeling problematic images in a semi-supervised fashion, as mentioned above.

As discussed later in Section \ref{Sec:labeling_results}, for images that are well represented in the training dataset, the classification results achieved by using the complete feature space do not significantly differ from those of the Softmax layer. That is, the representation of the feature space by the Softmax layer is as good as the complete feature space itself. However, if we consider problematic images selected by $f_a(\hat{p}_j,t)$, the classification accuracy improves drastically when using the complete feature space. On the other hand, classifying on the feature space is computationally expensive and unsuitable for online applications. As a result, ResNet50 should still be used for online classifications, while the feature space is used offline and only for images where the Softmax layer has failed.

As mentioned above, images selected by $f_a(\hat{p}_j,t)$ are not properly represented in the current training set $\mathcal{T}_q$. Hence, the idea is to use these images for a later training. To this end, a semi-supervised labeling process generates a label $l'_j$ --- also called pseudo-label \cite{dong-hyun_lee_pseudo-label_2013} --- for each such image $\mathbf{x}_j$ based on the whole feature space.

In order to calculate distances inside the feature space, each feature is normalized using $L2$ and denoted by $\mathbf{f}'_j$. For each class $c_n\in\mathcal{C}$, $M$ anchor points $a^n_m$, with  $1\leq n\leq N$ and $1\leq m \leq M$, are generated by $k$-means clustering on the normalized feature space $\mathcal{F}'_q$. The probability of $\mathbf{f}'_j$ to belong to class $c_n$ is calculated using soft voting:
\begin{eqnarray}
    \label{eq:SoftVoting}
    p'^{n}_j=\frac{\sum_{m=1}^{M} e^{-\gamma\|\mathbf{f}'_j-a_{m}^n\|_2^2}}{\sum_{z=1}^{N}\left(\sum_{m=1}^{M} e^{-\gamma\|\mathbf{f}'_j-a_{m}^z\|_2^2}\right)},
\end{eqnarray}
where $\gamma$ is the parameter controlling the {\em softness} of the label assignment, i.e., how much influence each anchor point has according to its distance from $\mathbf{f}'_j$ \cite{cui_fine-grained_2016}. Finally, the class $c_n$ with the highest confidence value is assigned to the label $l'_j=c_n \Leftrightarrow \hat{p}'_j=p'^n_j=\,\max(p'^1_j,\cdots,p'^N_j)$, where $\hat{p}'_j$ denotes the confidence values obtained by classifying in the complete feature space, as opposed to the confident value $\hat{p}_j$ obtained from the  Softmax layer. The labeling process forms a set $\mathcal{S}$ containing each selected image $\mathbf{x}_j$, its assigned label $l'_j$ and its feature vector $\mathbf{f}'_j$:
\begin{eqnarray}
    \label{eq:S}
    \mathcal{S}=\{(\mathbf{x}_j,l'_j,\mathbf{f}'_j), \forall\, \mathbf{x}_j\, |\, \hat{p}_j<t\}.
\end{eqnarray}
Once the set $\mathcal{S}$ reaches a given size $|\mathcal{S}|$, it is passed to the incremental learning process.
\subsection{Incremental Learning Process}
\label{sec:incremental_learning}

\noindent As stated above, images that could not be decided during the classification process are separated by the acquisition function $f_a(\hat{p}_j,t)$ and labeled by the semi-supervised labeling process leading to $S$ as per \eqref{eq:S}. When $S$ becomes sufficiently large, it is incorporated into the classification model by our incremental learning process.

Since not all $N$ classes may be represented with a similar number of examples in $\mathcal{S}$ or some classes may not be represented at all, fine-tuning can potentially lead to overfitting and catastrophic forgetting \cite{goodfellow_empirical_2013,chen_fine-tuning_2017}. To avoid this, $\mathcal{S}$ is {\em balanced} with images from $\mathcal{T}_q$. In particular, random images of each class from $\mathcal{T}_q$ are appended to  $\mathcal{S}$ until reaching a minimum number of $Q$ images per class. This results in a balanced set $\mathcal{S}'$ used to fine-tune the Softmax layer. This is performed offline using a copy of the current classification model $\mathcal{M}_q$.

This process allows evolving from $\mathcal{M}_q$ to $\mathcal{M}_{q+1}$, which already considers problematic images in $S$. By incorporating images and labels from $\mathcal{S}$ into the training dataset $\mathcal{T}_q$, this latter evolves to $\mathcal{T}_{q+1}$ and its size grows from $|\mathcal{T}_q|$ to $|\mathcal{T}_q|+|\mathcal{S}|$. The feature space is also updated from $\mathcal{F}_q$ to $\mathcal{F}_{q+1}$ by adding the features vectors $\mathbf{f}'_j \in \mathcal{S}$. Finally, $\mathcal{S}$ is emptied and the current $\mathcal{M}_q$ is replaced by the new one $\mathcal{M}_{q+1}$.

\section{\uppercase{Experiments and Results}}
\label{sec:experimet}

\noindent As already mentioned, the results reported in this work are based on Resnet50 \cite{he_deep_2015}, which has been pre-trainned with ImageNet \cite{deng_imagenet:_2009}. To test and validate our system, a dataset with four different classes ($N=4$): mug, bottle, bowl and chair was generated. To this end, we have extracted all RoIs (Region of Interest) from the Open Images Dataset \cite{openimages} that are labeled with one of the mentioned classes and have an area of at least $16,\!384$ pixels. As a result, we obtain a dataset of $19,\!207$ images. This set is then separated in two disjoint sets: the training dataset $\mathcal{T}_0$ and a validation dataset $\mathcal{V}$, where $|\mathcal{T}_0|=15,\!366$, $|\mathcal{V}|=3841$.% and, clearly, $\mathcal{T}\cap\mathcal{V}=\emptyset$.

The training dataset $\mathcal{T}_0$ is used to fine-tune the Softmax layer. The training is performed using a batch size of $128$, the Stochastic Gradient Descent optimizer, a learning rate of $0.0005$, a momentum of $0.9$, a maximum of $100$ epochs and early-stopping. After training  the resulting model $\mathcal{M}_0$ is used to generate the feature space $\mathcal{F}_0$.% that is required by the proposed approach.

Then the model $\mathcal{M}_0$ is tested on all images $\mathbf{x}_j \in \mathcal{V}$. The acquisition function $f_a(\hat{p}_j,t)$ --- see again \eqref{eq:Acquisition} --- is used to divide $\mathcal{V}$ in two disjoint subsets: $\mathcal{V}_k=\{(\mathbf{x}_j,l_j) \in \mathcal{V} \,|\, \hat{p}_j\geq t\}$ and $\mathcal{V}_{u}=\{(\mathbf{x}_j,l_j) \in \mathcal{V}\, |\, \hat{p}_j<t\}$. A threshold value of $t=0.9$ for classification confidence leads to $|\mathcal{V}_k|=3054$ and $|\mathcal{V}_{u}|=787$.

The classification of the images in $\mathcal{V}_k$ is considered correct ($\hat{p}_j\geq 0.9$) and we conclude that the system cannot learn more from them. On the other hand, $\mathcal{V}_{u}$ is formed by the images $\mathbf{x}_j$ that could not be classified with sufficient confidence. Clearly, these images contain information that the current model does not know and, hence, can be used to improve the system.

\subsection{Classifying and Labeling on the Feature Space}
\label{Sec:labeling_results}

\begin{figure}[h!]
    \centering
    \begin{subfigure}{0.45\columnwidth}
        \includegraphics[width=\columnwidth]{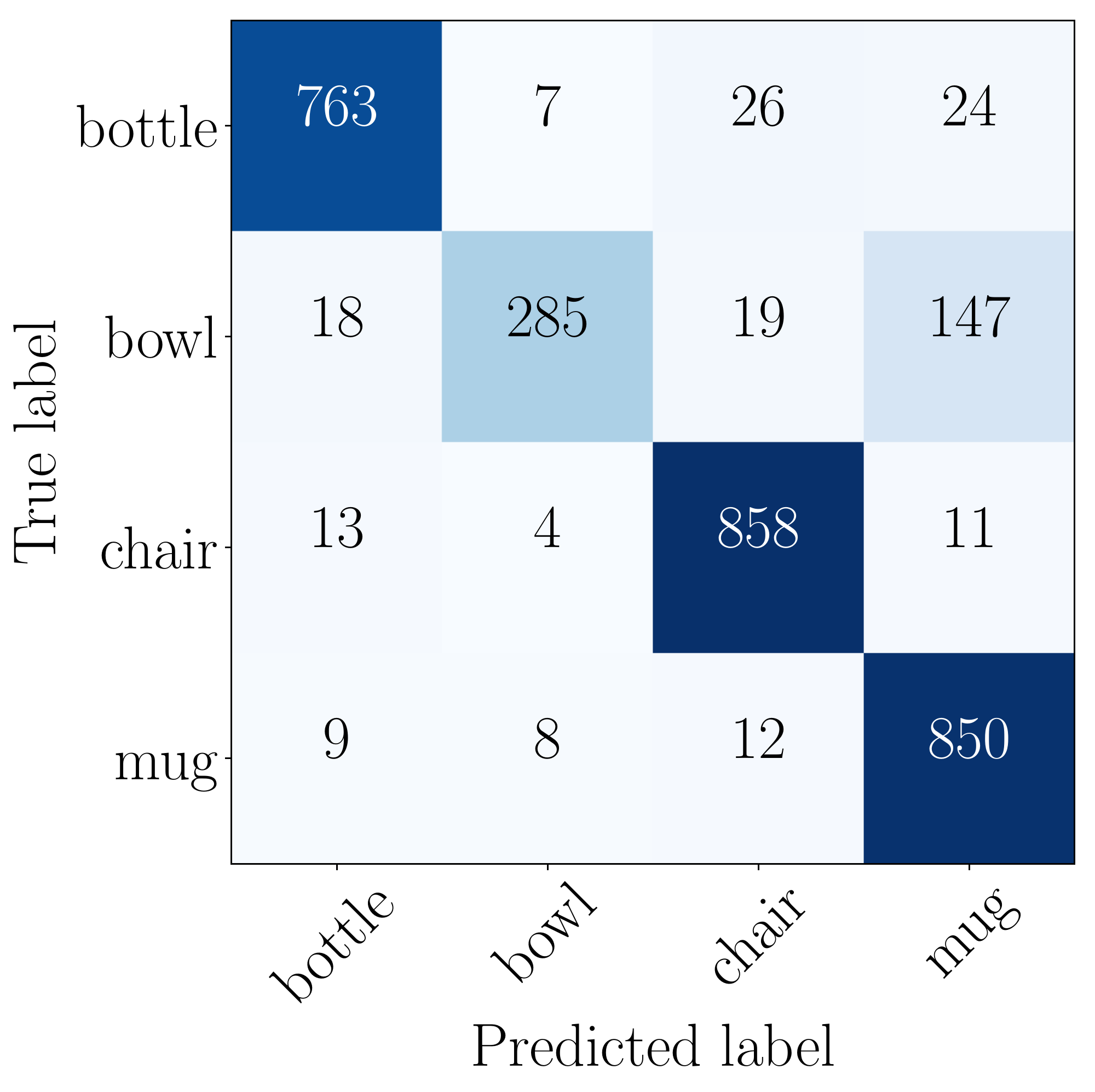}
        \caption{}
        \label{fig:labeling_SL_Vk}
    \end{subfigure}
    \begin{subfigure}{0.45\columnwidth}
        \includegraphics[width=\columnwidth]{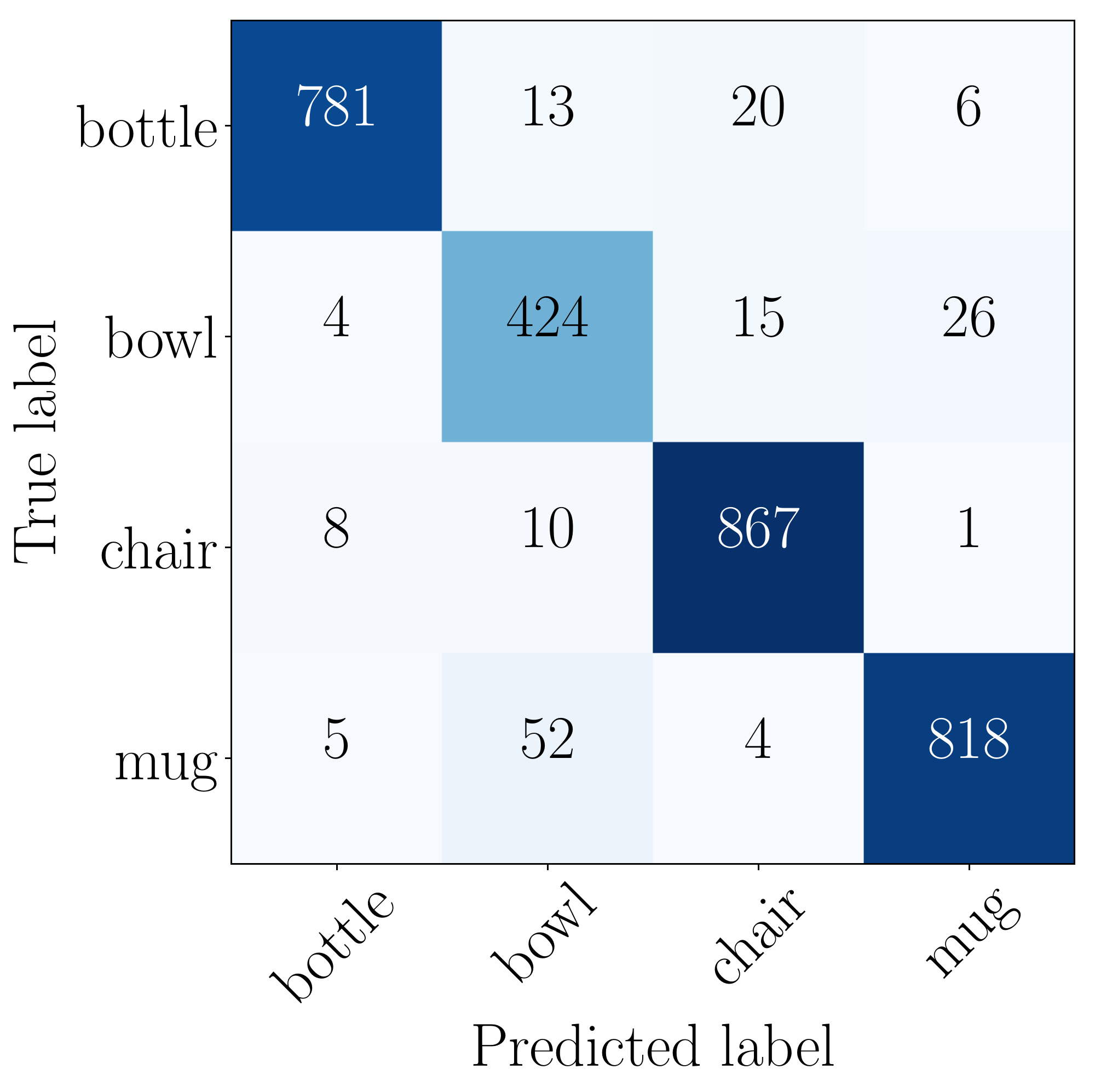}
        \caption{}
        \label{fig:Incremental_FS_Vk}
    \end{subfigure}
    \par\medskip
    \begin{subfigure}{0.45\columnwidth}
        \includegraphics[width=\columnwidth]{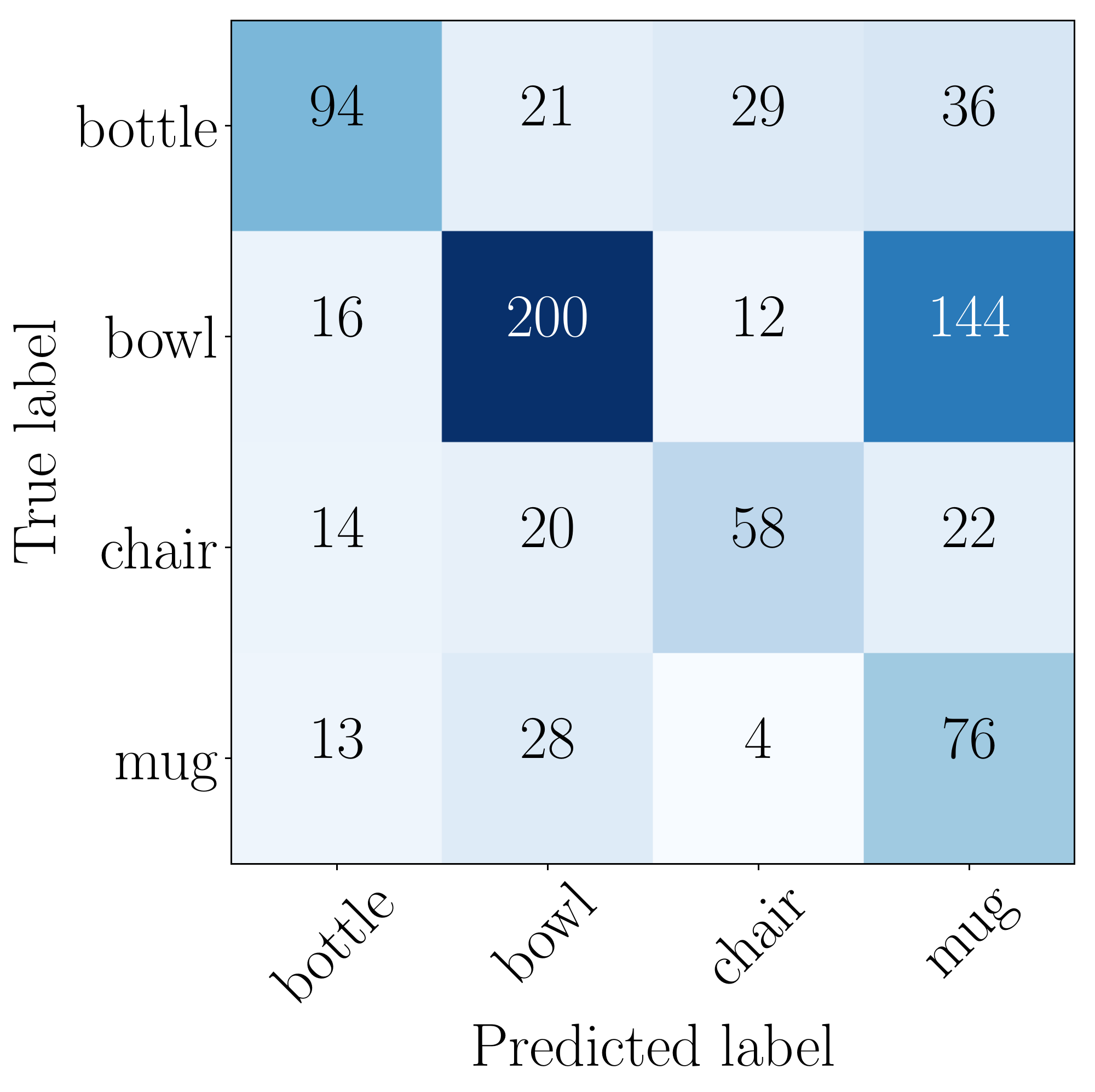}
        \caption{}
        \label{fig:Incremental_SL_Vu}
    \end{subfigure}
    \begin{subfigure}{0.45\columnwidth}
        \includegraphics[width=\columnwidth]{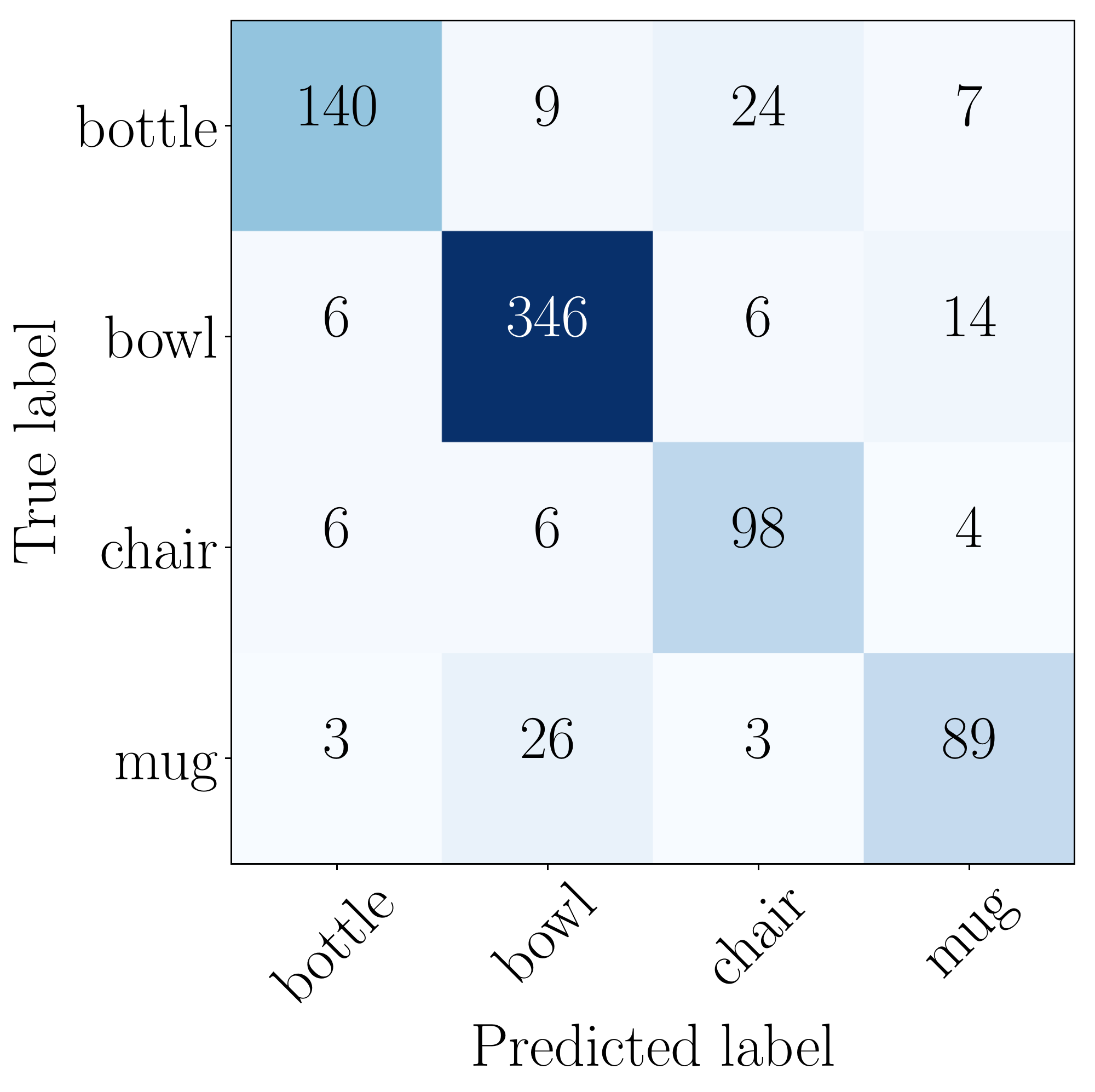}
        \caption{}
        \label{fig:Incremental_FS_Vu}
    \end{subfigure}
    \par\medskip

    \caption {Confusion matrices: (a) and (b) Softmax layer and feature space classification on $\mathcal{V}_{k}$, (c) and (d) Softmax layer and feature space classification on $\mathcal{V}_{u}$.} \label{fig:CM_labeling}
\end{figure}

\noindent In this section, we evaluate the efficiency of the feature space to classify images and specially to label problematic images. First, we use $\mathcal{V}_k$ to validate the feature space as a classifier by comparing its performance with that of the Softmax layer. This experiment is based on the feature space $\mathcal{F}_0$ (obtained from the original training dataset $\mathcal{T}_0$), where we consider $k = 10$, $M = 10$ and $\lambda = 1.5$ in \eqref{eq:SoftVoting}, and the corresponding model $\mathcal{M}_0$. As shown in Fig.~\ref{fig:labeling_SL_Vk}, $\mathcal{M}_0$ achieves an accuracy of $0.9$, whereas $\mathcal{F}_0$ reaches a slightly better accuracy of $0.94$ as per Fig.~\ref{fig:Incremental_FS_Vk}. This similar performance results from the fact that  $\mathcal{M}_0$ is already efficient at describing the images in $\mathcal{V}_k$ and, hence, using $\mathcal{F}_0$ does not considerably improve the classification accuracy.

To evaluate the feature space as a semi-supervised labeler, we now use $\mathcal{V}_u$. $\mathcal{V}_{u}$ contains problematic images selected by the acquisition function for $t=0.9$. This time, the Softmax layer yields a low accuracy of only $0.54$ as shown in Fig.~\ref{fig:Incremental_SL_Vu}. On the contrary, a classification using the complete feature space increases the accuracy to $0.85$ as shown in Fig.~\ref{fig:Incremental_FS_Vu}.%(and therefore the labeling based on it)
%
\begin{comment}
\begin{figure}[h]
    \centering
    \includegraphics[width=\columnwidth]{Figures/matrix.pdf}
    \caption{Confusion matrices labeling}
    \label{fig:CM_labeling}
\end{figure}
\end{comment}
%

This result confirms our hypothesis from Sec.\ref{sec:Labeling_Process} and validates using the complete feature space for labeling problematic images. As in any semi-supervised method, there is some label noise, which is about $15\%$ in our experiments. In the next section, we evaluate the incremental learning process with respect to robustness against label noise.

\subsection{Incremental Learning}
\label{sec:Incremental Learning}

\noindent To evaluate the incremental learning process, we split the set of {\em problematic} images $\mathcal{V}_{u}$ by randomly selecting pictures into two disjoint sets: $\mathcal{V}_{learn}$ and $\mathcal{V}_{test}$, with $|\mathcal{V}_{learn}|=472$ (i.e., around $3\over 5$ of the images in $\mathcal{V}_u$) and $|\mathcal{V}_{test}|=315$ (i.e., $|\mathcal{V}_{test}|=|\mathcal{V}_u|-|\mathcal{V}_{learn}|$).

$\mathcal{V}_{learn}$ is employed to investigate different $\mathcal{S}$ as defined in \eqref{eq:S} with $0<|\mathcal{S}|\leq|\mathcal{V}_{learn}|$, while $\mathcal{V}_{test}$ and $\mathcal{V}_k$ are used to evaluate results. Testing on $\mathcal{V}_{test}$ allows us to evaluate how much the system improves after learning from $\mathcal{S}$, i.e., how good it starts recognizing problematic images. On the other hand, testing on $\mathcal{V}_{k}$ helps evaluating how much the systems worsens after learning from $\mathcal{S}$, i.e., whether it stops recognizing some known images.

To avoid {\em catastrophic forgetting}, the Softmax layer must be fine-tuned with a {\em well-balanced} image set. To this end, similar to \cite{chen_fine-tuning_2017}, we fix the number of images for each class to be equal to $Q$. That is, we enforced $|\mathcal{S}^n|=Q$ for each class $c_n$, where $\mathcal{S}^n=\{\mathbf{x}_j\,|\,l'_j=c_n\}$, $\mathcal{S}^n \subset \mathcal{S}$ and $Q>\max_{\forall n}(|\mathcal{S}^n|)$ hold. The $Q-|\mathcal{S}^n|$ additional images necessary to balance each class in $\mathcal{S}$ are randomly selected from $\mathcal{T}_q$. This is also valid for classes that may not be represented in $\mathcal{S}$ ($\mathcal{S}^n=\emptyset$), where all $Q$ images are extracted from $\mathcal{T}_q$.

\textbf{Finding an optimum size for $\mathcal{S}$.} To investigate how the size of $\mathcal{S}$ affects the classification accuracy, we generate a sequence of sets by randomly selecting $b$ images from $\mathcal{V}_{learn}$. The generated $\floor{|\mathcal{V}_{lean}|/b}$ sets are then successively used to fine-tune the Softmax layer. The resulting classification models from $\mathcal{M}_0$ (before starting with incremental learning) to $\mathcal{M}_{\floor{|\mathcal{V}_{lean}|/b}}$ (after fine-tuning with all images in $V_{learn}$) are tested on $\mathcal{V}_{test}$ and $\mathcal{V}_k$.

\begin{figure}[h!]
    \centering
    \begin{subfigure}{\columnwidth}
        \includegraphics[trim=0 0 0 100,clip,width=\columnwidth]{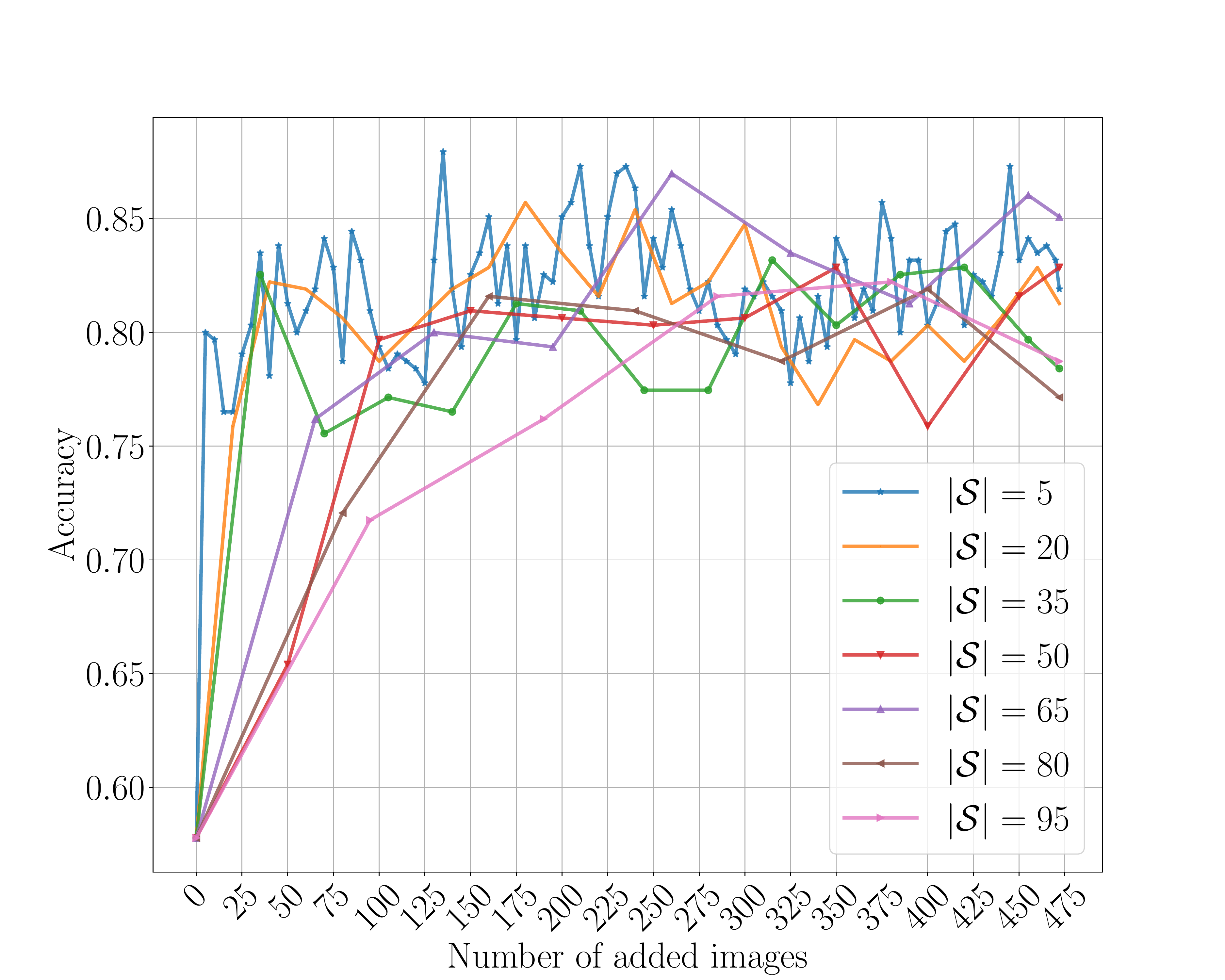}
        \caption{Test on $\mathcal{V}_{test}$}
        \vspace{0.3cm}
        \label{fig:Incremental_Ssize_Test}
    \end{subfigure}
    \begin{subfigure}{\columnwidth}
        \includegraphics[trim=0 0 0 100,clip,width=\columnwidth]{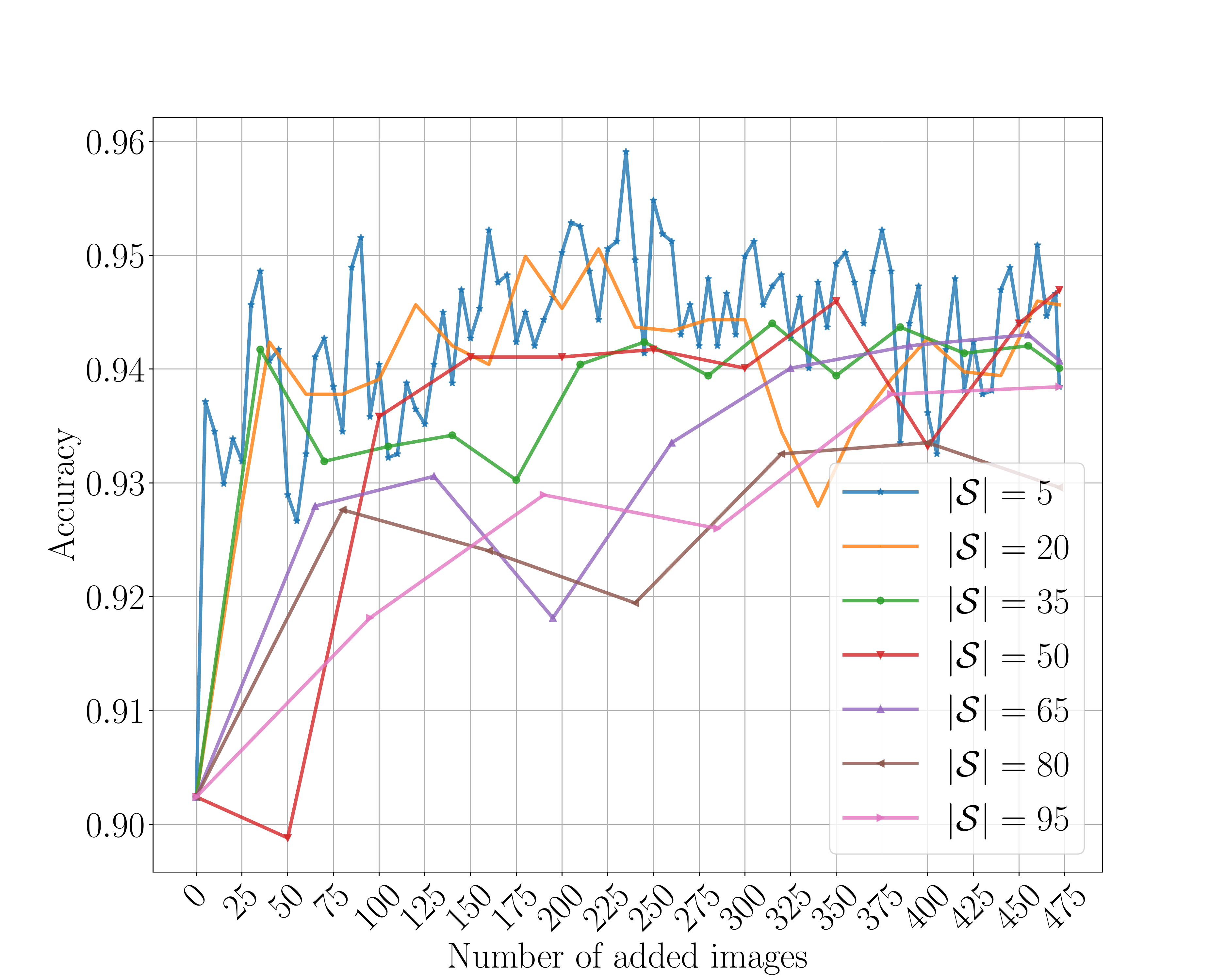}
        \caption{Test on $\mathcal{V}_k$}
        \label{fig:Incremental_Ssize_Vk}
    \end{subfigure}
    \caption {Incremental learning process for $|\mathcal{S}|=\{5,20,35,50,65,80,95\}$ and $Q=100$.}\label{fig:Incremental_Ssize}
\end{figure}

We vary $b$ from $5$ to $95$ in steps of $15$, which results in $|\mathcal{S}|=\{5,20,35,50,65,80,95\}$. Since we randomly select images from $\mathcal{V}_{learn}$ to form $\mathcal{S}$ and from $\mathcal{T}_q$ to balance $\mathcal{S}$, every run of this experiment leads to slightly different results. Hence, to reduce randomization effects, Fig.~\ref{fig:Incremental_Ssize_Test} and Fig.~\ref{fig:Incremental_Ssize_Vk} show the average result over three independent runs of the experiment.

Figure \ref{fig:Incremental_Ssize_Test} shows that the network is able to learn already from $|\mathcal{S}|=5$ onward considering $Q=100$. A small $|\mathcal{S}|$ allows us to update of the classification model faster, since less problematic images need to be collected for an update. In addition, since the updated classification model is expected to perform better, less images will be considered as problematic next time, which reduces the number of iterations. For these reasons, we select $|\mathcal{S}|=5$ for the next experiments. For $|S|<5$, results start worsening, since $S$ does not provide enough new information anymore.

Note that, for a fixed $Q$, the larger the size of $\mathcal{S}$ the lower the percentage of {\em known} images in the balanced $\mathcal{S}$. As we can see in Fig.~\ref{fig:Incremental_Ssize_Vk}, for $|\mathcal{S}|\geq 65$, the classification accuracy decreases as $Q=100$ is to small and we start overfitting for the images in $\mathcal{S}$. However, at the end of the learning process, when all images of $\mathcal{V}_{learn}$ have been incorporated, the difference in accuracy is lower than $2\%$ for all values of $|\mathcal{S}|$.

\begin{figure}[h!]
    \centering
    \includegraphics[trim=0 0 0 20,clip,width=\columnwidth]{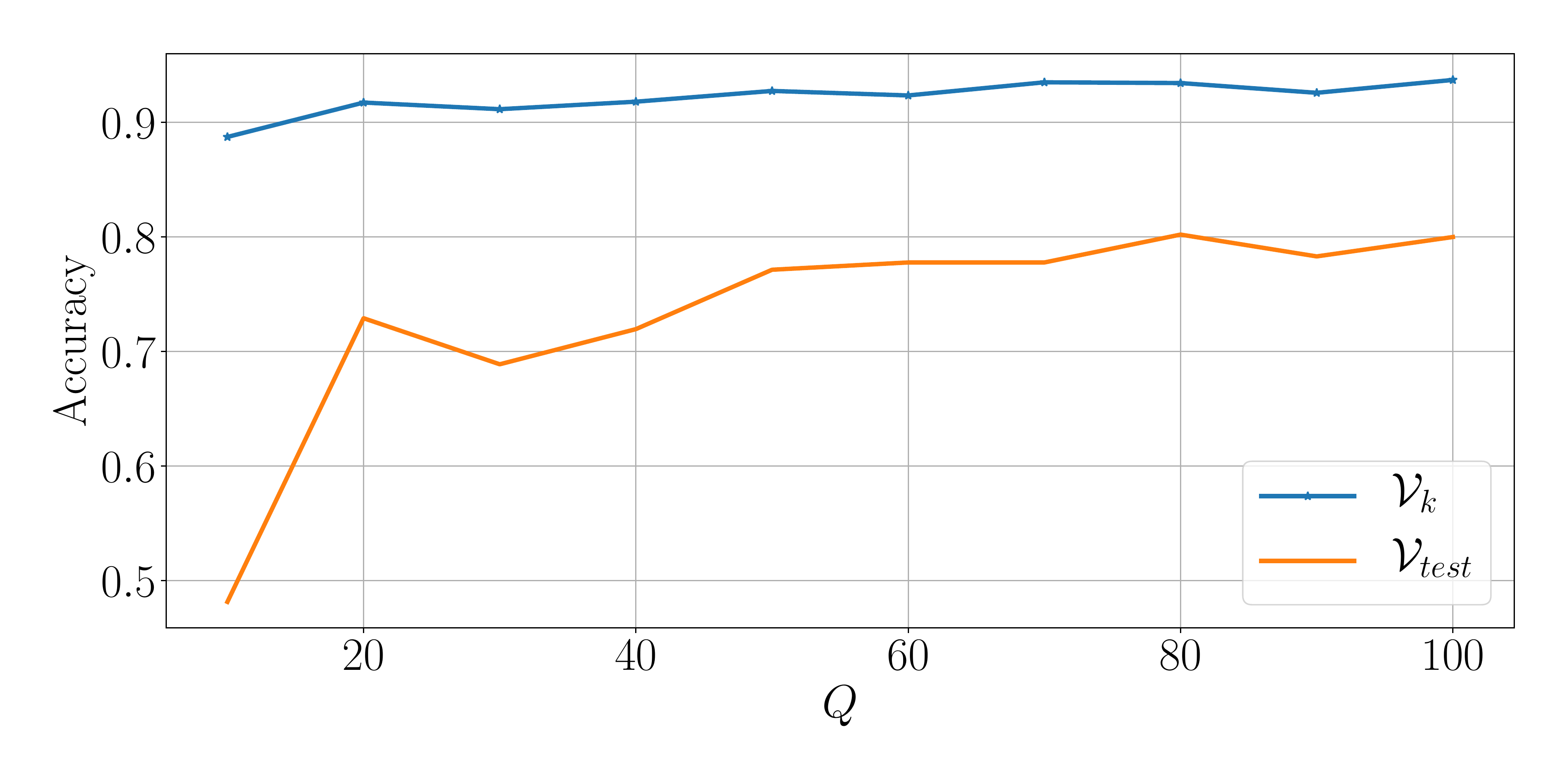}
    \caption {Accuracy of $\mathcal{M}_1$ for $|\mathcal{S}|=5$ and $10 \leq Q \leq 100$.}\label{fig:Q_S5}
\end{figure}

\textbf{Finding an optimum $Q$.} In the previous experiment, we have fixed $Q$ to $100$ in order to study different values of $|\mathcal{S}|$. However, the optimum $Q$ depends on the value of $|\mathcal{S}|$. Figure \ref{fig:Q_S5} shows the influence of different values of $Q$ on the classification accuracy for $|\mathcal{S}|=5$. Each point of this plot represents the average accuracy reached by $\mathcal{M}_1$ (i.e., after the first incremental learning step) for 3 independent runs of the experiment when varying $Q$ from $10$ bis $100$. The accuracy for $\mathcal{M}_0$ on $\mathcal{V}_{test}$ and $\mathcal{V}_{k}$ (see Fig.~\ref{fig:Incremental_Ssize_Test} and \ref{fig:Incremental_Ssize_Vk}) is of $0.574$ and $0.902$ respectively. Consequently, as shown in Fig.~\ref{fig:Q_S5}, $\mathcal{M}_1$ starts outperforming $\mathcal{M}_0$ from $Q\geq 20$ onward.

\begin{figure}
    \centering
    \includegraphics[trim=0 0 0 100,clip,width=\columnwidth]{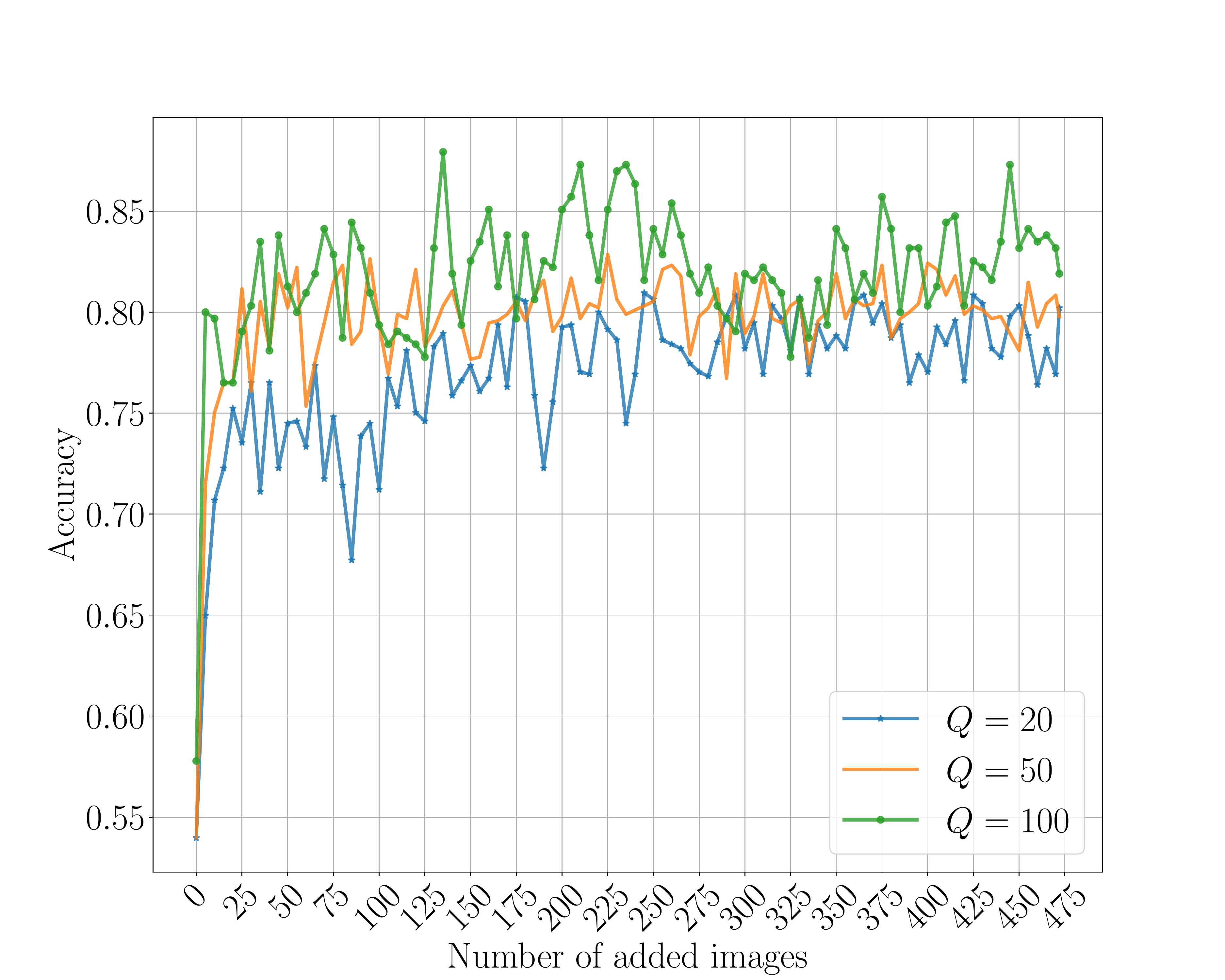}
    \caption {Incremental learning process with $|\mathcal{S}|=5$ for $Q=\{20, 50, 100\}$ tested on $\mathcal{V}_{test}$.}
    \label{fig:Incremental_learning_SQ}
\end{figure}

\begin{figure}
    \centering
    \includegraphics[trim=0 0 0 35,clip,width=\columnwidth]{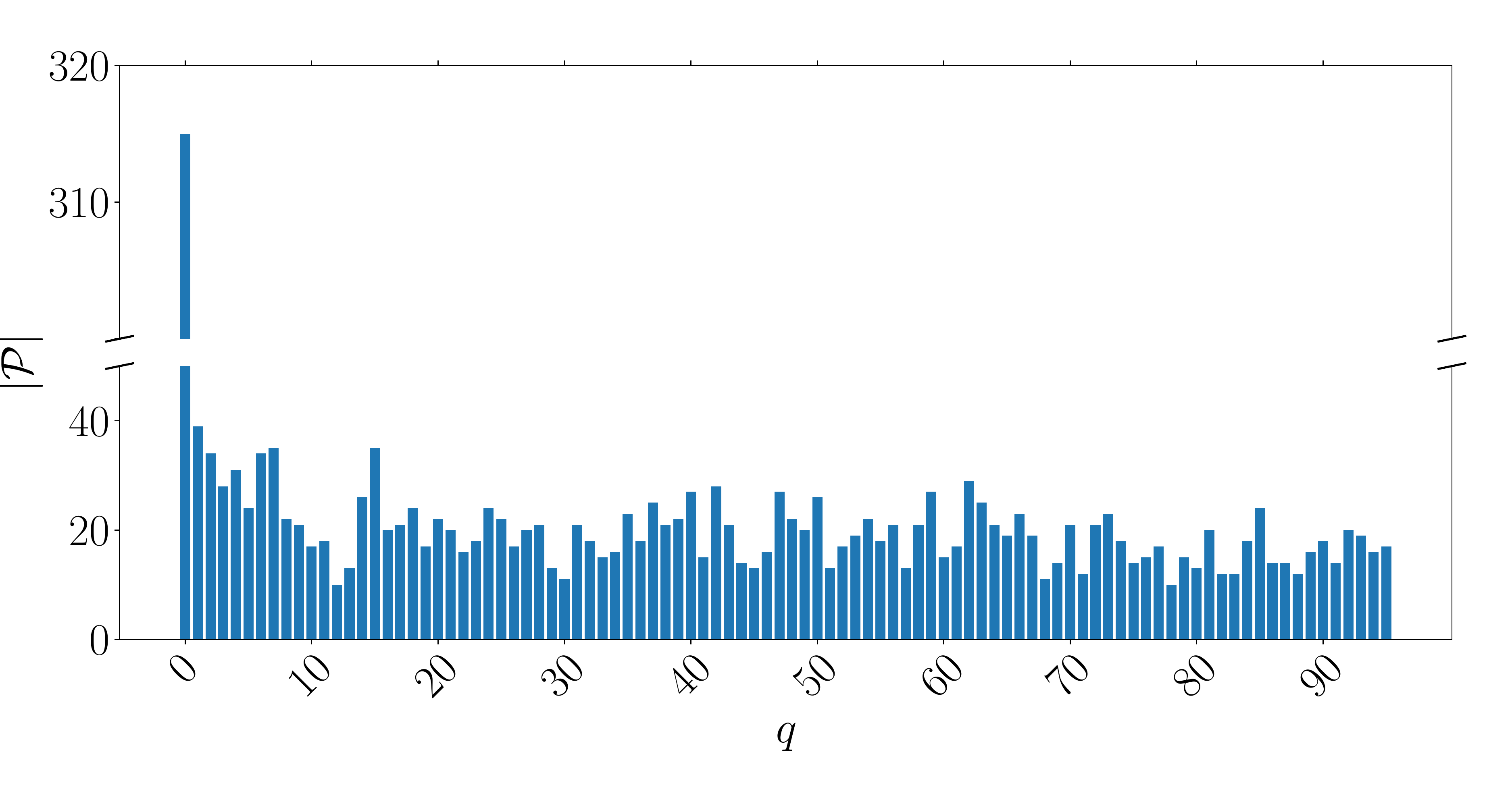}
    \caption{$|\mathcal{P}|$ number of images in $\mathcal{V}_{test}$ classified with $\hat{p}_j<0.9$ for each model $\mathcal{M}_q$ where $0\leq q \leq \lceil|\mathcal{V}_{test}|/|\mathcal{S}|\rceil$.}
    \label{fig:Incremental_learning}
\end{figure}

\begin{figure}[t]
    \begin{center}
        \includegraphics[trim=0 0 0 100,clip,width=\linewidth]{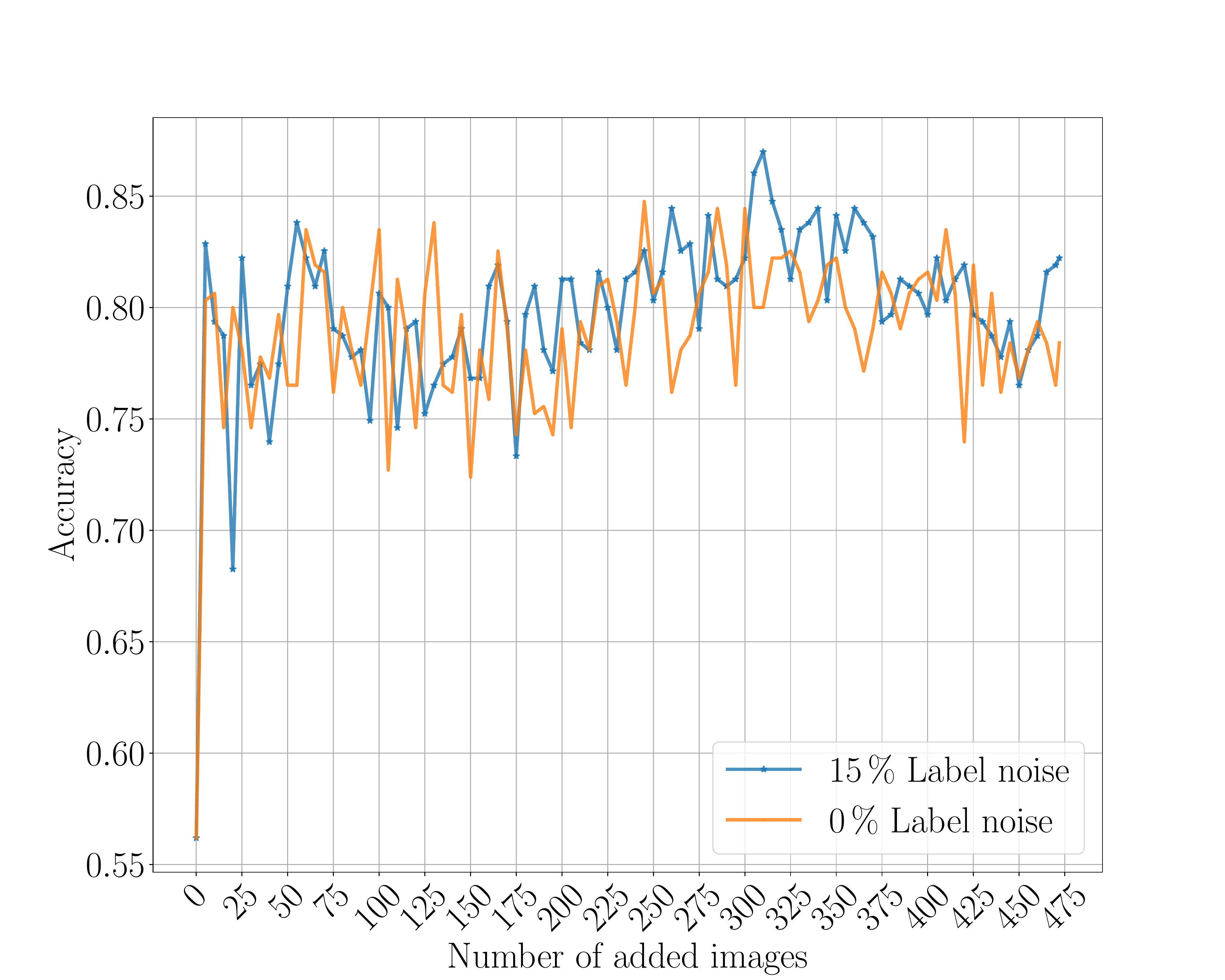}
    \end{center}
    \caption {Incremental learning process for $|\mathcal{S}|=5$, $Q=100$ with and without label noise tested on $\mathcal{V}_{test}$.}\label{fig:Incremental_LN}
\end{figure}

Finally Fig.~\ref{fig:Incremental_learning_SQ} shows the average classification accuracy on $\mathcal{V}_{test}$ for 3 independent experiments along the whole learning process considering $|\mathcal{S}|=5$ and $Q=\{20,50,100\}$. The accuracy achieved at the end of the incremental learning process is almost the same for all three values of $Q$, where $Q=100$ shows the best results for $|\mathcal{S}|=5$. Same conclusions are obtained by testing on $\mathcal{V}_k$.
\textbf{Analyzing classification confidence.} As mentioned above, after each incremental learning step, the number of images classified with confidence values that are below $0.9$ decreases. This latter results in reducing the number of incremental learning iterations. To visualize this effect, Fig.~\ref{fig:Incremental_learning} shows the number of images $|\mathcal{P}|$ in $\mathcal{V}_{test}$ that are classified with a confidence $\hat{p}_j<0.9$ after every update of the classification model (with $|\mathcal{S}|=5$ and $Q=100$), i.e., $\mathcal{P}=\{\mathbf{x}_j\in\mathcal{V}_{test}\,|\,\hat{p}_j<0.9\}$. Again, we repeat this experiment $3$ times and average results. It is remarkable that the number of problematic images already falls from $315$ to $39$ after the first update. In other words, selecting larger $\mathcal{S}$ delays any update of the model ending up collecting images with redundant information.
\textbf{Evaluation of label noise.} As discussed in Sec.~\ref{Sec:labeling_results}, the proposed semi-supervised labeling has an error rate of approximately $15\%$ leading to wrong labels.  Figure \ref{fig:Incremental_LN} illustrates how this label noise impacts incremental learning by plotting the classification results on $\mathcal{V}_{test}$ with and without label noise. We again repeated this experiment $3$ times with $|\mathcal{S}|=5$ and $Q=100$ concluding that this amount of label noise has a negligible impact on accuracy along the incremental learning process. Same results are obtained by testing on $\mathcal{V}_k$.
\section{\uppercase{Conclusions and future work}}
\label{Sec:Conclusion}

\noindent In this paper, we proposed an approach for a CNN-based semi-supervised incremental learning that efficiently handles new instances. Our approach leverages the feature space generated from the training dataset of the CNN to automatically label problematic images. Even though there is some label noise (around $15\%$), we show that classification results considerably improve when updating the CNN's classification model with the information contained in these images. To avoid catastrophic forgetting we proposed a combination of partial rehearsal and early stopping.

Our results indicate an improvement of around $40\%$ more correctly detected new instances with respect to the case of no incremental learning. Moreover, there is also an improvement of $4\%$ in the classification accuracy of known images. That is, by learning from problematic images, the CNN is also able to correct false classification results, that were not detected by the acquisition function because of their high confidence values. Finally, as future work, we plan to extend our system to the case where new object classes need to be learned.

\section*{\uppercase{Acknowledgements}}
This work is funded by the European Regional Development Fund (ERDF) under the grant number 100-241-945.
{\small
\bibliographystyle{apalike}
\bibliography{paper}

\begin{thebibliography}{}

\bibitem[Cui et~al., 2016]{cui_fine-grained_2016}
Cui, Y., Zhou, F., Lin, Y., and Belongie, S. (2016).
\newblock Fine-{Grained} {Categorization} and {Dataset} {Bootstrapping} {Using}
  {Deep} {Metric} {Learning} with {Humans} in the {Loop}.
\newblock In {\em 2016 {IEEE} {Conference} on {Computer} {Vision} and {Pattern}
  {Recognition} ({CVPR})}, pages 1153--1162, Las Vegas, NV, USA. IEEE.

\bibitem[Deng et~al., 2009]{deng_imagenet:_2009}
Deng, J., Dong, W., Socher, R., Li, L.-J., {Kai Li}, and {Li Fei-Fei} (2009).
\newblock {ImageNet}: {A} large-scale hierarchical image database.
\newblock In {\em 2009 {IEEE} {Conference} on {Computer} {Vision} and {Pattern}
  {Recognition}}, pages 248--255, Miami, FL.

\bibitem[{Dong-Hyun Lee}, 2013]{dong-hyun_lee_pseudo-label_2013}
{Dong-Hyun Lee} (2013).
\newblock Pseudo-{Label} : {The} {Simple} and {Efficient} {Semi}-{Supervised}
  {Learning} {Method} for {Deep} {Neural} {Networks}.
\newblock In {\em {ICML} 2013 {Workshop}: {Challenges} in {Representation}
  {Learning} ({WREPL})}, Atlanta, Georgia, USA.

\bibitem[Enguehard et~al., 2019]{enguehard_semi-supervised_2019}
Enguehard, J., O'Halloran, P., and Gholipour, A. (2019).
\newblock Semi-{Supervised} {Learning} {With} {Deep} {Embedded} {Clustering}
  for {Image} {Classification} and {Segmentation}.
\newblock {\em IEEE Access}, 7:11093--11104.

\bibitem[Gal et~al., 2017]{gal_deep_2017}
Gal, Y., Islam, R., and Ghahramani, Z. (2017).
\newblock Deep {Bayesian} {Active} {Learning} with {Image} {Data}.
\newblock In {\em {ICML}'17 {Proceedings} of the 34th {International}
  {Conference} on {Machine} {Learning}}, Sydney, Australia.

\bibitem[Goodfellow et~al., 2013]{goodfellow_empirical_2013}
Goodfellow, I.~J., Mirza, M., Xiao, D., Courville, A., and Bengio, Y. (2013).
\newblock An {Empirical} {Investigation} of {Catastrophic} {Forgetting} in
  {Gradient}-{Based} {Neural} {Networks}.
\newblock {\em arXiv:1312.6211 [cs, stat]}.
\newblock arXiv: 1312.6211.

\bibitem[Han et~al., 2018]{han_advanced_2018}
Han, J., Zhang, D., Cheng, G., Liu, N., and Xu, D. (2018).
\newblock Advanced {Deep}-{Learning} {Techniques} for {Salient} and
  {Category}-{Specific} {Object} {Detection}: {A} {Survey}.
\newblock {\em IEEE Signal Processing Magazine}, 35(1):84--100.

\bibitem[Hayes et~al., 2018]{hayes_memory_2018}
Hayes, T.~L., Cahill, N.~D., and Kanan, C. (2018).
\newblock Memory {Efficient} {Experience} {Replay} for {Streaming} {Learning}.
\newblock {\em arXiv:1809.05922 [cs, stat]}.
\newblock arXiv: 1809.05922.

\bibitem[He et~al., 2016]{he_deep_2015}
He, K., Zhang, X., Ren, S., and Sun, J. (2016).
\newblock Deep {Residual} {Learning} for {Image} {Recognition}.
\newblock In {\em 2016 {IEEE} {Conference} on {Computer} {Vision} and {Pattern}
  {Recognition} ({CVPR})}, pages 770--778, Las Vegas, NV, USA.

\bibitem[Kemker et~al., 2018]{kemker_measuring_2017}
Kemker, R., McClure, M., Abitino, A., Hayes, T., and Kanan, C. (2018).
\newblock Measuring {Catastrophic} {Forgetting} in {Neural} {Networks}.
\newblock New Orleans, Louisiana, USA.

\bibitem[Kirkpatrick et~al., 2017]{kirkpatrick_overcoming_2016}
Kirkpatrick, J., Pascanu, R., Rabinowitz, N., Veness, J., Desjardins, G., Rusu,
  A.~A., Milan, K., Quan, J., Ramalho, T., Grabska-Barwinska, A., Hassabis, D.,
  Clopath, C., Kumaran, D., and Hadsell, R. (2017).
\newblock Overcoming catastrophic forgetting in neural networks.
\newblock {\em Proceedings of the National Academy of Sciences},
  114(13):3521--3526.

\bibitem[Krasin et~al., 2017]{openimages}
Krasin, I., Duerig, T., Alldrin, N., Ferrari, V., Abu-El-Haija, S., Kuznetsova,
  A., Rom, H., Uijlings, J., Popov, S., Veit, A., Belongie, S., Gomes, V.,
  Gupta, A., Sun, C., Chechik, G., Cai, D., Feng, Z., Narayanan, D., and
  Murphy, K. (2017).
\newblock Openimages: A public dataset for large-scale multi-label and
  multi-class image classification.
\newblock {\em Dataset available from https://github.com/openimages}.

\bibitem[Käding et~al., 2017]{chen_fine-tuning_2017}
Käding, C., Rodner, E., Freytag, A., and Denzler, J. (2017).
\newblock Fine-{Tuning} {Deep} {Neural} {Networks} in {Continuous} {Learning}
  {Scenarios}.
\newblock In Chen, C.-S., Lu, J., and Ma, K.-K., editors, {\em Computer
  {Vision} – {ACCV} 2016 {Workshops}}, volume 10118, pages 588--605. Springer
  International Publishing, Cham.

\bibitem[Li and Hoiem, 2018]{li_learning_2016}
Li, Z. and Hoiem, D. (2018).
\newblock Learning without {Forgetting}.
\newblock {\em IEEE Transactions on Pattern Analysis and Machine Intelligence},
  40(12):2935--2947.

\bibitem[Lin et~al., 2014]{fleet_microsoft_2014}
Lin, T.-Y., Maire, M., Belongie, S., Hays, J., Perona, P., Ramanan, D.,
  Dollár, P., and Zitnick, C.~L. (2014).
\newblock Microsoft {COCO}: {Common} {Objects} in {Context}.
\newblock In Fleet, D., Pajdla, T., Schiele, B., and Tuytelaars, T., editors,
  {\em Computer {Vision} – {ECCV} 2014}, volume 8693, pages 740--755.
  Springer International Publishing, Cham.

\bibitem[Lomonaco and Maltoni, 2017]{lomonaco_core50:_2017}
Lomonaco, V. and Maltoni, D. (2017).
\newblock {CORe}50: a {New} {Dataset} and {Benchmark} for {Continuous} {Object}
  {Recognition}.
\newblock In {\em Proceedings of the 1st {Annual} {Conference} on {Robot}
  {Learning}}, California, USA.

\bibitem[Maltoni and Lomonaco, 2019]{maltoni_continuous_2018}
Maltoni, D. and Lomonaco, V. (2019).
\newblock Continuous learning in single-incremental-task scenarios.
\newblock {\em Neural Networks}, 116:56--73.

\bibitem[McCloskey and Cohen, 1989]{mccloskey_catastrophic_1989}
McCloskey, M. and Cohen, N.~J. (1989).
\newblock Catastrophic {Interference} in {Connectionist} {Networks}: {The}
  {Sequential} {Learning} {Problem}.
\newblock In {\em Psychology of {Learning} and {Motivation}}, volume~24, pages
  109--165. Elsevier.

\bibitem[Parisi et~al., 2019]{parisi_continual_2019}
Parisi, G.~I., Kemker, R., Part, J.~L., Kanan, C., and Wermter, S. (2019).
\newblock Continual lifelong learning with neural networks: {A} review.
\newblock {\em Neural Networks}, 113:54--71.

\bibitem[Rasmus et~al., 2015]{rasmus_semi-supervised_2015}
Rasmus, A., Valpola, H., Honkala, M., Berglund, M., and Raiko, T. (2015).
\newblock Semi-{Supervised} {Learning} with {Ladder} {Networks}.
\newblock In {\em {NIPS}'15 {Proceedings} of the 28th {International}
  {Conference} on {Neural} {Information} {Processing} {Systems}}, volume~2,
  pages 3546--3554, Montreal, Canada. MIT Press.

\bibitem[Rebuffi et~al., 2017]{rebuffi_icarl:_2016}
Rebuffi, S.-A., Kolesnikov, A., Sperl, G., and Lampert, C.~H. (2017).
\newblock {iCaRL}: {Incremental} {Classifier} and {Representation} {Learning}.
\newblock In {\em The {IEEE} {Conference} on {Computer} {Vision} and {Pattern}
  {Recognition} ({CVPR})}, Honolulu, Hawaii.

\bibitem[Russakovsky et~al., 2015]{russakovsky_imagenet_2015}
Russakovsky, O., Deng, J., Su, H., Krause, J., Satheesh, S., Ma, S., Huang, Z.,
  Karpathy, A., Khosla, A., Bernstein, M., Berg, A.~C., and Fei-Fei, L. (2015).
\newblock {ImageNet} {Large} {Scale} {Visual} {Recognition} {Challenge}.
\newblock {\em International Journal of Computer Vision}, 115(3):211--252.

\bibitem[Rusu et~al., 2016]{rusu_progressive_2016}
Rusu, A.~A., Rabinowitz, N.~C., Desjardins, G., Soyer, H., Kirkpatrick, J.,
  Kavukcuoglu, K., Pascanu, R., and Hadsell, R. (2016).
\newblock Progressive {Neural} {Networks}.
\newblock {\em arXiv:1606.04671 [cs]}.
\newblock arXiv: 1606.04671.

\end{thebibliography}
}

\end{document}